\documentclass{article}

\usepackage[preprint,nonatbib]{neurips_2024}

\usepackage[utf8]{inputenc} 
\usepackage[T1]{fontenc}    
\usepackage{hyperref}       
\usepackage{url}            
\usepackage{booktabs}       
\usepackage{amsfonts}       
\usepackage{nicefrac}       
\usepackage{microtype}      
\usepackage[dvipsnames]{xcolor}         
\usepackage{amsmath}
\usepackage{makecell}
\usepackage{graphicx}
\usepackage{multirow}
\usepackage{tabu}
\usepackage{pifont}

\newcommand{\parsection}[1]{\textbf{#1} }

\title{\model{}: Dynamic Multi-Object Scene Generation from Monocular Videos}

\author{
    \parbox{\linewidth}{\centering
        Wen-Hsuan Chu$^{\dagger}$,\hspace{0.1cm}
        Lei Ke$^{\dagger}$,\hspace{0.1cm} 
        Katerina Fragkiadaki \\
        {\normalfont Carnegie Mellon University} \\
        {\tt \small{\{wenhsuac,leik,katef\}@cs.cmu.edu}}
    } \\ \\
    \textcolor{magenta}{\url{https://dreamscene4d.github.io/}}
}

\begin{document}

\newcommand{\model}{DreamScene4D}
\newcommand{\cmark}{\ding{51}}
\newcommand{\xmark}{\ding{55}}

\maketitle
\let\thefootnote\relax\footnotetext{$^{\dagger}$Equal contribution}

\begin{abstract}
View-predictive generative models provide strong priors for lifting object-centric images and videos into 3D and 4D through rendering and score distillation objectives. 
A question then remains: what about lifting complete multi-object dynamic scenes? 
There are two challenges in this direction: First, rendering error gradients are often insufficient to recover fast object motion, and second,  view predictive generative models work much better for objects than whole scenes, so, score distillation objectives cannot currently be applied at the scene level directly. 
We present \model{}, the first approach to generate 3D dynamic scenes of multiple objects from monocular videos via $360^\circ$ novel view synthesis.
Our key insight is a ``\textit{decompose-recompose}'' approach that factorizes the video scene into the background and object tracks, while also factorizing object motion into 3 components: object-centric deformation, object-to-world-frame transformation, and camera motion.
Such decomposition permits rendering error gradients and object view-predictive models to recover object 3D completions and deformations while bounding box tracks guide the large object movements in the scene.
We show extensive results on challenging DAVIS, Kubric, and self-captured videos with quantitative comparisons and a user preference study. 
Besides 4D scene generation,~\model{} obtains accurate 2D persistent point track by projecting the inferred 3D trajectories to 2D.
We will release our code and hope our work will stimulate more research on fine-grained 4D understanding from videos.
\end{abstract}

\section{Introduction}
\label{sec:intro}
\begin{figure}[t]
    \vspace{-0.1in}
    \centering
    \includegraphics[width=1.0\textwidth]{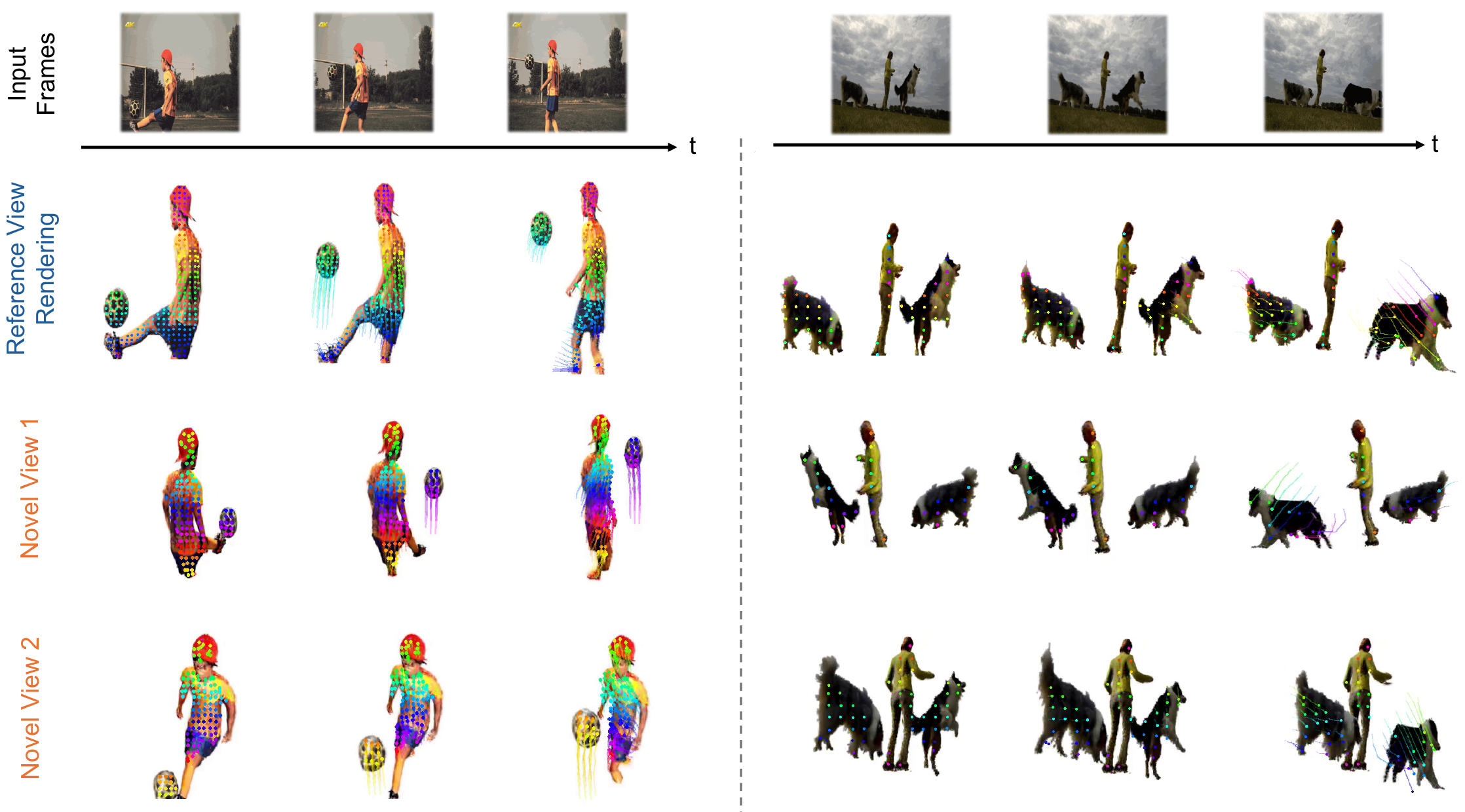}
    \vspace{-0.25in}    \caption{\textbf{\model{}} extends video-to-4D generation to multi-object videos with fast motion. We present rendered images and the corresponding motions from diverse viewpoints at different timesteps using real-world DAVIS~\cite{pont20172017} videos with multiple objects and large motions.}
    \label{fig:teaser}
    \vspace{-0.1in}
\end{figure}
Videos are the result of entities moving and interacting in 3D space and over time, captured from a moving camera.
Inferring the dynamic 4D scene from video projections in terms of complete 3D object reconstructions and their 3D motions across seen and unseen camera views is a challenging problem in computer vision.
It has multiple important applications, such as 3D object and scene state tracking for robot perception~\cite{heiden2022inferring,peng2018sfv}, action recognition, visual imitation, digital content creation/simulation, and augmented reality.

Video-to-4D is a highly under-constrained problem since multiple 4D generation hypotheses project to the same video observations.
Existing 4D reconstruction works~\cite{pumarola2021d, park2021nerfies, li2022neural, lombardi2019neural, cao2023hexplane, liu2020neural} 
mainly focus on the visible part of the scene contained in the video by learning a differentiable 3D representation that is often a neural field~\cite{mildenhall2020nerf} or a set of 3D Gaussians~\cite{kerbl20233d} with temporal deformation. 
\textit{What about the unobserved views of the dynamic 3D scene}? Existing 4D generation works utilize generative models to constrain the appearance of objects in unseen views through score distillation losses. Text-to-4D~\cite{singer2023text,bahmani20234d,xu2024comp4d,ling2023align,bahmani2024tc4d} or image-to-4D~\cite{zhao2023animate124,ren2023dreamgaussian4d,yang2024beyond,zheng2023unified} setups take a single text prompt or image as input to create a 4D object.
Several works~\cite{jiang2023consistent4d,gao2024gaussianflow,ren2023dreamgaussian4d,pan2024fast,yin20234dgen,zeng2024stag4d} explore the video-to-4D setup, but these methods predominantly focus on videos containing a single object with minor 3D deformations, where the object deforms in place without large motion in the 3D scene.
This focus arises because current generative models perform significantly better at predicting novel views of individual objects~\cite{liu2023zero} than multi-object scenes.
Consequently, score distillation objectives for 3D object lifting are difficult to apply directly at a scene level.
Also, optimization difficulty arises when neural fields or 3D Gaussians are trained to model large temporal deformations directly.
This limits their practical real-world usage where input videos depicting real-world complex scenes containing multiple dynamic objects with fast motions, as illustrated in Figure~\ref{fig:exp_vis}.

In this paper, we propose~\model{}, the first video-to-4D scene generation approach to produce realistic 4D scene representation from a complex multi-object video with large object motion or deformation. 
To $360^\circ$ synthesize novel views for multiple objects of the scene,~\model{} proposes a ``decompose-recompose'' strategy. A video is first decomposed into objects and the background scene, where each is completed across occlusions and viewpoints, then recomposed to estimate relative scales and rigid object-to-world transformations in each frame using monocular depth guidance, so all objects are placed back in a common coordinate system. 

To handle fast-moving objects,~\model{} factorizes the 3D motion of the static object Gaussians into 3 components: 1) camera motion, 2) object-centric deformations, and 3) an object-centric to world frame transformation. 
This factorization greatly improves the stability of the motion optimization process by leveraging powerful object trackers~\cite{cheng2022xmem} to handle large motions and allowing view-predictive generative models to receive object-centric inputs that are in distribution.
The camera motion is estimated by re-rendering the static background Gaussians to match the video frames. 

We show the view renderings at various timesteps and diverse viewpoints of \model{} using challenging monocular videos from DAVIS~\cite{pont20172017} in Figure~\ref{fig:teaser}.
\model{} achieves significant improvements compared to the existing SOTA video-to-4D generation approaches~\cite{ren2023dreamgaussian4d,jiang2023consistent4d} on DAVIS, Kubric~\cite{greff2022kubric}, and our self-captured videos with fast moving objects (Figures~\ref{fig:exp_vis}).
To evaluate the quality of the learned Gaussian motions, we measure the 2D endpoint error (EPE) of the inferred 3D motion trajectories across occlusions and show that our approach produces accurate and persistent point trajectories in both visible views and synthesized novel views.

\section{Related Work}
\label{sec:related}
\parsection{Video-to-4D Reconstruction}
Dynamic 3D reconstruction extends static 3D reconstruction to dynamic scenes with the goal of 3D lifting the visible parts of the video. 
Dynamic NeRF-based methods~\cite{pumarola2021d,park2021nerfies,li2022neural,liu2023robust,busching2023flowibr} extend NeRF~\cite{mildenhall2020nerf} to dynamic scenes, typically using grid or voxel-based representations \cite{lombardi2019neural, cao2023hexplane, liu2020neural}, or learning a deformation field~\cite{cao2023hexplane,fridovich2023k} that models the dynamic portions of an object or scene. Dynamic Gaussian Splatting~\cite{luiten2023dynamic} extends 3D Gaussian Splatting~\cite{kerbl20233d}, where scenes are represented as 4D Gaussians and show faster convergence than NeRF-based approaches.
However, these 4D scene reconstruction works~\cite{luiten2023dynamic,wu20234d,yang2023deformable3dgs} typically take videos where the camera has a large number of multi-view angles, instead of a general monocular video input. This necessitates precise calibration of multiple cameras and constrains their potential real-world applicability.
Different from these works~\cite{park2021hypernerf,wu20234d} on mostly reconstructing the visible regions of the dynamic scene, \model{} can $360^\circ$ synthesize novel views for multiple objects of the scene, including the unobserved regions in the video.  

\parsection{Video-to-4D Generation}
In contrast to 4D reconstruction works, this line of research is most related by attempting to complete and 3D reconstruct a video scene across both visible and unseen (virtual) viewpoints.
Existing text to image to 4D generation works~\cite{ren2023dreamgaussian4d,jiang2023consistent4d,gao2024gaussianflow,pan2024fast,yin20234dgen,zeng2024stag4d}
typically use score distillation sampling (SDS)~\cite{poole2022dreamfusion} to supply constraints in unseen viewpoints in order to synthesize full 4D representations of objects from single text \cite{singer2023text,bahmani20234d,ling2023align,bahmani2024tc4d}, image \cite{yang2024beyond,zheng2023unified}, or a combination of both~\cite{zhao2023animate124} prompts.
They first map the text prompt or image prompt to a synthetic video, then lift the latter using deformable 3D differentiable NeRFs~\cite{li2022neural} or set of Gaussians~\cite{wu20234d} representation.
Existing video-to-4D generation works~\cite{ren2023dreamgaussian4d,jiang2023consistent4d,gao2024gaussianflow,pan2024fast,yin20234dgen,zeng2024stag4d} usually simplify the input video by assuming a non-occluded and slow-moving object while real-world videos with multiple dynamic objects inevitably contain occlusions. 
Owing to our proposed scene decoupling and motion factorization schemes, \model{} is the first approach to generate complicated 4D scenes and synthesize their arbitrary novel views by taking real-world videos of multi-object scenes.

\section{Approach}
\label{method:overview}
To generate dynamic 4D scenes of multiple objects from a monocular video input, we propose \model{}, which takes Gaussian Splatting~\cite{kerbl20233d,wu20234d} as the 4D scene representation and leverages powerful foundation models to generalize to diverse zero-shot settings. 

\subsection{Background: Generative 3D Gaussian Splatting}
\label{sec:background}
Gaussian Splatting~\cite{kerbl20233d} represents a scene with a set of 3D Gaussians. 
Each Gaussian is defined by its centroid, scale, rotation, opacity, and color, represented as spherical harmonics (SH) coefficients.

\parsection{Generative 3D Gaussian Splatting via Score Distillation Sampling} 
Score Distillation Sampling (SDS)~\cite{poole2022dreamfusion} is widely used for text-to-3D or image-to-3D tasks by leveraging a diffusion prior for optimizing 3D Gaussians to synthesize novel views.
For 3D object generation, DreamGaussian~\cite{tang2023dreamgaussian} uses Zero-1-to-3 \cite{liu2023zero}, which takes a reference view and a relative camera pose as input and generates plausible images for the target viewpoint, for single frame 2D-to-3D lifting.
The 3D Gaussians of the input reference view $I_1$ are optimized by a rendering loss and an SDS loss~\cite{poole2022dreamfusion}:

\begin{equation}
    \nabla_\phi \mathcal{L}_\text{SDS}^\text{t}=\mathbb{E}_{t, \tau, \epsilon, p}\left[w(\tau)\left(\epsilon_\theta\left(\hat{I}^p_t ; \tau, I_1, p\right)-\epsilon\right) \frac{\partial \hat{I}^p_t}{\partial \phi}\right],
    \label{eq:sds_loss_image}
\end{equation}
where $t$ is the timestep indices, $w( \tau )$ is a weighting function for denoising timestep $\tau$, $\phi\left(\cdot\right)$ represents the Gaussian rendering function, $\hat{I}^p_t$ is the rendered image, $\epsilon_\theta \left( \cdot \right)$ is the predicted noise from Zero-1-to-3, and $\epsilon$ is the added noise. We take the superscript $p$ to represent an arbitrary camera pose.

\subsection{DreamScene4D}

\begin{figure}[t]
  \begin{minipage}[c]{0.67\textwidth}
    \includegraphics[width=1.0\textwidth]{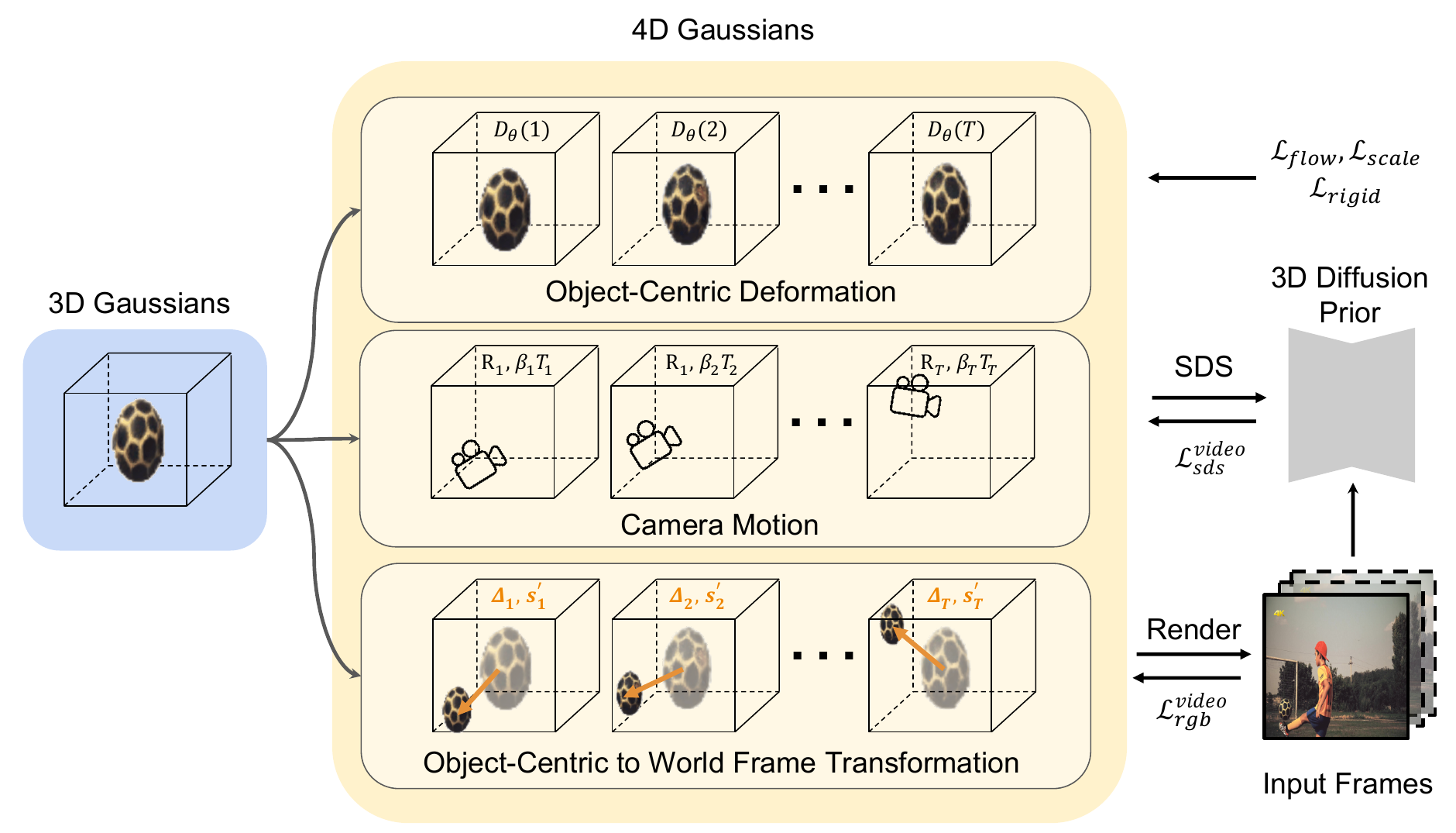}
  \end{minipage}\hfill
  \begin{minipage}[c]{0.3\textwidth}
    \caption{\textbf{3D Motion Factorization.} The 3D motion is decomposed into 3 components: 1) the object-centric deformation, 2) the camera motion, and 3) the object-centric to-world frame transformation. After optimization, they can be composed to form the original object motion observed in the video.}
    \label{fig:method_motion_fact}
  \end{minipage}
  \vspace{-0.25in}
\end{figure}

We propose a ``\textit{decompose-recompose}'' principle to handle complex multi-object scenes.
As in Figure~\ref{fig:method_high_level},
given a monocular video of multiple objects, we first segment and track~\cite{ren2024grounded,sam_hq,cheng2022xmem,chu2023zero} each 2D object and recover the appearance of the occluded regions (Section \ref{method:scene_decompose}). 
Next, we decompose the scene into multiple amodal objects and use SDS~\cite{poole2022dreamfusion} with diffusion priors to obtain a 3D Gaussian representation for each object (Section~\ref{sec:crop_lift}). 
To handle large object motions, we optimize the deformation of 3D Gaussians under various constraints and factorize the motion into three components (Figure~\ref{fig:method_motion_fact}): the object-centric motion, an object-centric to world frame transformation, and the camera motion (Section~\ref{method:motion_fact}). This greatly improves the stability and quality of the Gaussian optimization and allows view-predictive image generative models to operate under in-distribution object-centric settings. 
Finally, we compose each individually optimized object to form a complete 4D scene representation using monocular depth guidance (Section \ref{method:composing}).

\subsubsection{Video Scene Decomposition}
\label{method:scene_decompose}

\begin{figure}[!t]
    \centering
    \includegraphics[width=1.0\textwidth]{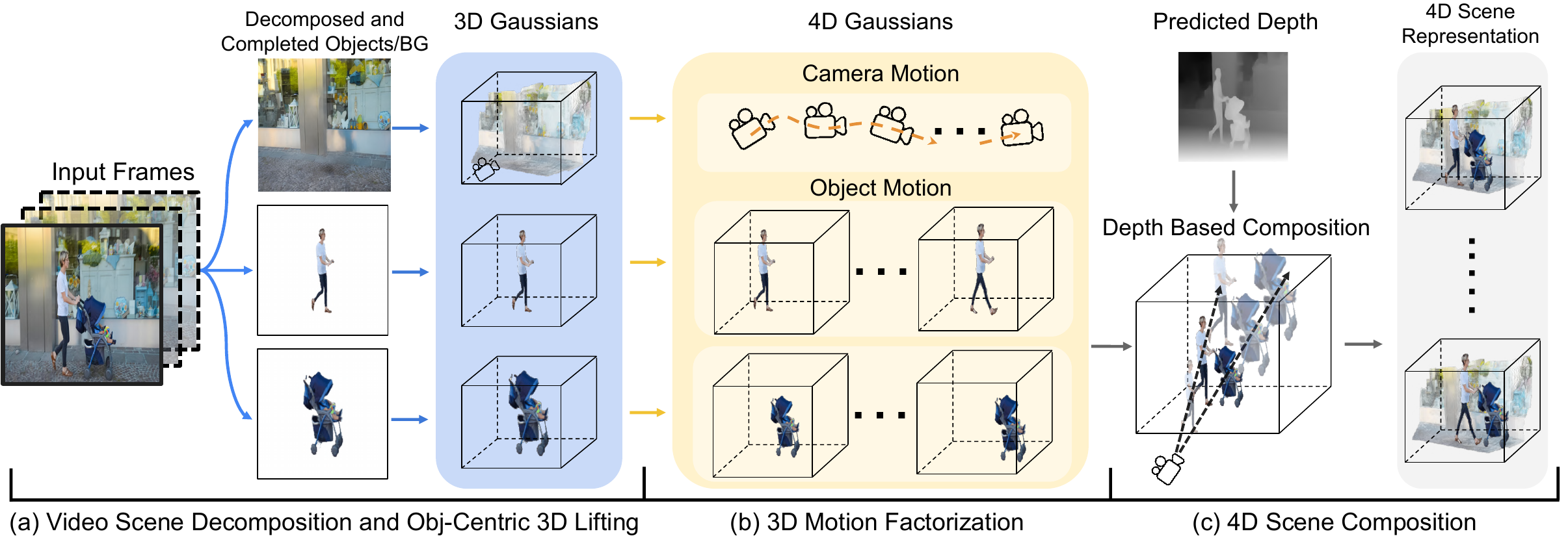}
    \caption{\textbf{Method overview for \model{}}: \textbf{(a)} We first \textit{decompose} and amodally complete each object and the background in the video sequence and use DreamGaussian~\cite{tang2023dreamgaussian} to obtain static 3D Gaussian representation. \textbf{(b)} Next, we factorize and optimize the motion of each object track independently, detailed in Figure~\ref{fig:method_motion_fact}. \textbf{(c)} Finally, we use the estimated monocular depth to \textit{recompose} the independently optimized 4D Gaussians into one unified coordinate frame.}
    \label{fig:method_high_level}
    \vspace{-1em}
\end{figure}

Instead of taking the video scene as a whole, we first adopt mask trackers~\cite{ren2024grounded,sam_hq,cheng2022xmem,chu2023zero} to segment and track objects in the monocular video when GT object masks are not provided. From the monocular video and associated object tracks, we amodally complete each object track before lifting it to 3D as in Figure~\ref{fig:method_high_level}. To achieve zero-shot object appearance recovery for occluded regions of individual object tracks, we build off of inpainting diffusion models~\cite{Rombach2022stablediffusion} and extend it to videos for amodal video completion. We provide the details of amodal video completion in the appendix.

 \subsubsection{Object-Centric 3D Lifting from World Frame} 
\label{sec:crop_lift}
After decomposing the scene into individual object tracks, we use Gaussians Splatting~\cite{kerbl20233d} with SDS loss~\cite{poole2022dreamfusion,tang2023dreamgaussian} to lift them to 3D. 
Since novel-view generative models~\cite{liu2023zero} trained on Objaverse~\cite{deitke2023objaverse} are inherently object-centric, we take a different manner to 3D lifting. Instead of directly using the first frame of the original video, where the object areas may be small and not centered, we create a new object-centric frame $\Tilde{I}_1$ by cropping the object using its bounding box and re-scaling it.
Then, we optimize the static 3D Gaussians with both the RGB rendering on $\Tilde{I}_1$ and the SDS loss~\cite{poole2022dreamfusion} in Eq.~\ref{eq:sds_loss_image}.

\subsubsection{Modeling Complex 3D Motions via Motion Factorization}
\label{method:motion_fact}

To estimate the motion of the first-frame lifted 3D Gaussians $\{G^{obj}_1\}$, one solution like DreamGaussian4D~\cite{ren2023dreamgaussian4d} is to model the object dynamics by optimizing the deformation of the 3D Gaussians directly in the world frame. 
However, this approach falls short in videos with large object motion, as the rendering loss yields minimal gradients until there is an overlap between the deformed Gaussians in the re-rendered frames and the objects in the video frames.
Large motions of thousands of 3D Gaussians also increase the training difficulty of the lightweight deformation network~\cite{fridovich2023k,cao2023hexplane}.

Thus, we propose to decompose the motion into three components and independently model them: \textbf{1)} object-centric motion, modeled using a learnable deformation network; \textbf{2)} the object-centric to world frame transformation, represented by a set of 3D displacements vectors and scaling factors; and \textbf{3)} camera motion, represented by a set of camera pose changes. Once optimized, the three components can be composed to form the object motion observed in the video.

\parsection{Object-Centric Motion Optimization}
\label{method:motion_obj_centric}
The deformation of the 3D Gaussians includes a set of learnable parameters for each Gaussian: \textbf{1)} a 3D position for each timestep $\mu_t = \left( \mu x_t, \mu y_t, \mu z_t \right)$, \textbf{2)} a 3D rotation for each timestep, represented by a quaternion $\mathcal{R}_t = \left( qw_t, qx_t, qy_t, qz_t \right)$, and \textbf{3)} a 3D scale for each timestep $s_t = \left( sx_t, sy_t, sz_t \right)$. The RGB (spherical harmonics) and opacity of the Gaussians are shared across all timesteps and copied from the first-frame 3D Gaussians.

To compute the 3D object motion in the object-centric frames,
we take the cropped and scaled objects in the individual frames $I_t$, forming a new set of frames $\Tilde{I}_t^r$ for each object. Following DreamGaussian4D~\cite{ren2023dreamgaussian4d}, we adopt a K-plane~\cite{fridovich2023k} based deformation network $D_{\theta} ( G^{obj}_1, t )$ to predict the 10-D deformation parameters ($\mu_t, R_t, s_t$) for each object per timestep. We denote the rendered image at timestep $t$ under the camera pose $p$ as $\hat{I}^p_t$, and optimize $D_{\theta}$ using the SDS loss in Eq.~\ref{eq:sds_loss_image}, as well as the rendering loss between $\hat{I}^r_t$ and $\Tilde{I}_t^r$ for each frame under the reference camera pose $r$.

Since 3D Gaussians can freely move within uniformly-colored regions without penalties, the rendering and SDS loss are often insufficient for capturing accurate motion, especially for regions with near-uniform colors. Thus, we additionally introduce a flow rendering loss $\mathcal{L}_{flow}$, which is the masked L1 loss between the rendered optical flow of the Gaussians and the flow predicted by an off-the-shelf optical flow estimator~\cite{xu2022gmflow}. 
The flow rendering loss only applies to the confident masked regions that pass a simple forward-backward flow consistency check.

\parsection{Physical Prior on Object-Centric Motion}
Object motion in the real world follows a set of physics laws, which can be used to constrain the Gaussian deformations further. 
For example, objects usually maintain a similar size in temporally neighboring frames. Thus, we incorporate a scale regularization loss $\mathcal{L}_\text{scale} = \frac{1}{T}\sum_{t=1}^{T} \left\| s_{t+1} - s_{t}  \right\|_1$, where we penalize large Gaussian scale changes.

To preserve the local rigidity during deformations, we apply a loss $\mathcal{L}_\text{rigid}$ to penalize changes to the relative 3D distance and orientation between neighboring Gaussians following \cite{luiten2023dynamic}.
We disallow pruning and densification of the Gaussians when optimizing for deformations like~\cite{ren2023dreamgaussian4d, luiten2023dynamic}.

\parsection{Object-to-world Frame Transformation}
\label{method:motion_global_warp}
We compute the translation $\Delta_t = \left( \Delta_{x, t},\, \Delta_{y, t},\, \Delta_{z, t} \right)$ and scaling factor $s'_t$ that warps the Gaussians from the object-centric frame to the world frame. The 2D bounding-box-based cropping and scaling (Sec~\ref{sec:crop_lift}) from the original frames to the object-centric frames can be represented as an affine warp, which we use to compute and initialize $\Delta_{x, t}$, $\Delta_{y, t}$, and $s'_t$ for each object in each frame. $\Delta_{z,t}$ is initialized to $0$. We then adopt the rendering loss on the original frames $I_t$ instead of center-cropped frames $\Tilde{I}_t$ to fine-tune $\Delta_t$ with a low learning rate.

To further improve the alignment between renderings and the video frames, it is essential to consider the perceptual parallax difference. This arises when altering the object's 3D position while maintaining a fixed camera perspective, resulting in subtle changes in rendered object parts. Thus, we compose the individually optimized motion components and jointly fine-tune the deformation network $D_{\theta}$ and affine displacement $\Delta_t$ using the rendering loss. This refinement process, conducted over a limited number of iterations, helps mitigate the parallax effect as shown in Figure~\ref{fig:supp_parallax} of the appendix.

\parsection{Camera Motion Estimation}
\label{method:motion_camera}
We leverage differentiable Gaussian Splatting rendering to jointly reconstruct the 3D static video background and estimate camera motions.
Taking multi-frame inpainted background images $I_t^\text{bg}$ as input, we first use an off-the-shelf algorithm~\cite{dust3r_cvpr24} to initialize the background Gaussians and relative camera rotation and translation~$\{R_t, T_t\}$ between frame $1$ and frame $t$. However, the camera motion can only be estimated up to an unknown scale \cite{ye2023decoupling} as there is no metric depth usage. Therefore, we also estimate a scaling term $\beta$ for ${T_t}$. Concretely, from the background Gaussians $G^{bg}$ and $\{R_t, T_t\}$, we find the $\beta$ that minimizes the rendering loss of the background in subsequent frames:
\begin{equation}
    \mathcal{L}_\text{bg} = \frac{1}{T}\sum_{t=1}^T \left\| I_t^\text{bg} - \phi \left( G^{bg},\, R_t,\, \beta_t T_t \right) \right\|_2,
\end{equation}
Empirically, optimizing a separate $\beta_t$ per frame~\cite{bian2023nope} yields better results by allowing the renderer to compensate for erroneous camera pose predictions.

\subsubsection{4D Scene Composition with Monocular Depth Guidance}
\label{method:composing}
Given the individually optimized 4D Gaussians, we recompose them into a unified coordinate frame to form a coherent 4D scene. 
As illustrated in Step (c) of Figure~\ref{fig:method_high_level}, this requires determining the depth and scale for each object along camera rays.

Concretely, we use an off-the-shelf depth estimator~\cite{depthanything} to compute the depth of each object and the background and exploit the relative depth relationships to guide the composition. 
We randomly pick an object as the ``reference'' object and estimate the relative depth scale $k$ between the reference object and all other objects. 
Then, the original positions $\mu_t^{\prime}$ and scales $s_t^{\prime}$ of the 3D Gaussians for the objects are scaled along the camera rays given this initialized scaling factor $k$: $\mu_t^{\prime} = \mathcal{C}^r - \left( \mathcal{C}^r - \mu_t \right) * k$ and $s_t^{\prime} = s_t * k$,
where $\mathcal{C}^r$ represents the position of the camera. 
Finally, we compose and render the depth map of the reference and scaled object, and minimize the affine-invariant L1 loss~\cite{depthanything, Ranftl2022} between the rendered and predicted depth map to optimize each object's scaling factor $k$:
\begin{equation}
    \mathcal{L}_{depth} =\frac{1}{H W} \sum_{i=1}^{H W} \left\| \hat{d}_i^* - \hat{d}_i \right\|_1, \,\,
    \hat{d}_i = \frac{d_i-t(d)}{\sigma(d)}.
    \label{eq:loss_depth}
\end{equation}
Here, $\hat{d}_i^*$ and $\hat{d}_i$ are the scaled and shifted versions of the rendered depth $d_i^*$ and predicted depth $d_i$.
$t(d)$ is defined as the reference object's median depth and $\sigma(d)$ is defined as the difference between the $90\%$ and $10\%$ quantile of the reference object. The two depth maps are normalized separately using their own $t(d)$ and $\sigma(d)$. Once we obtain the scaling factor $k$ for each object, we can easily place and re-compose the individual objects in a common coordinate frame. The Gaussians can then be rendered jointly to form a scene-level 4D representation.

\section{Experiments}
\label{exp}
\begin{figure}[t]
    \centering
    \vspace{-0.15in}
    \includegraphics[width=1.0\textwidth]{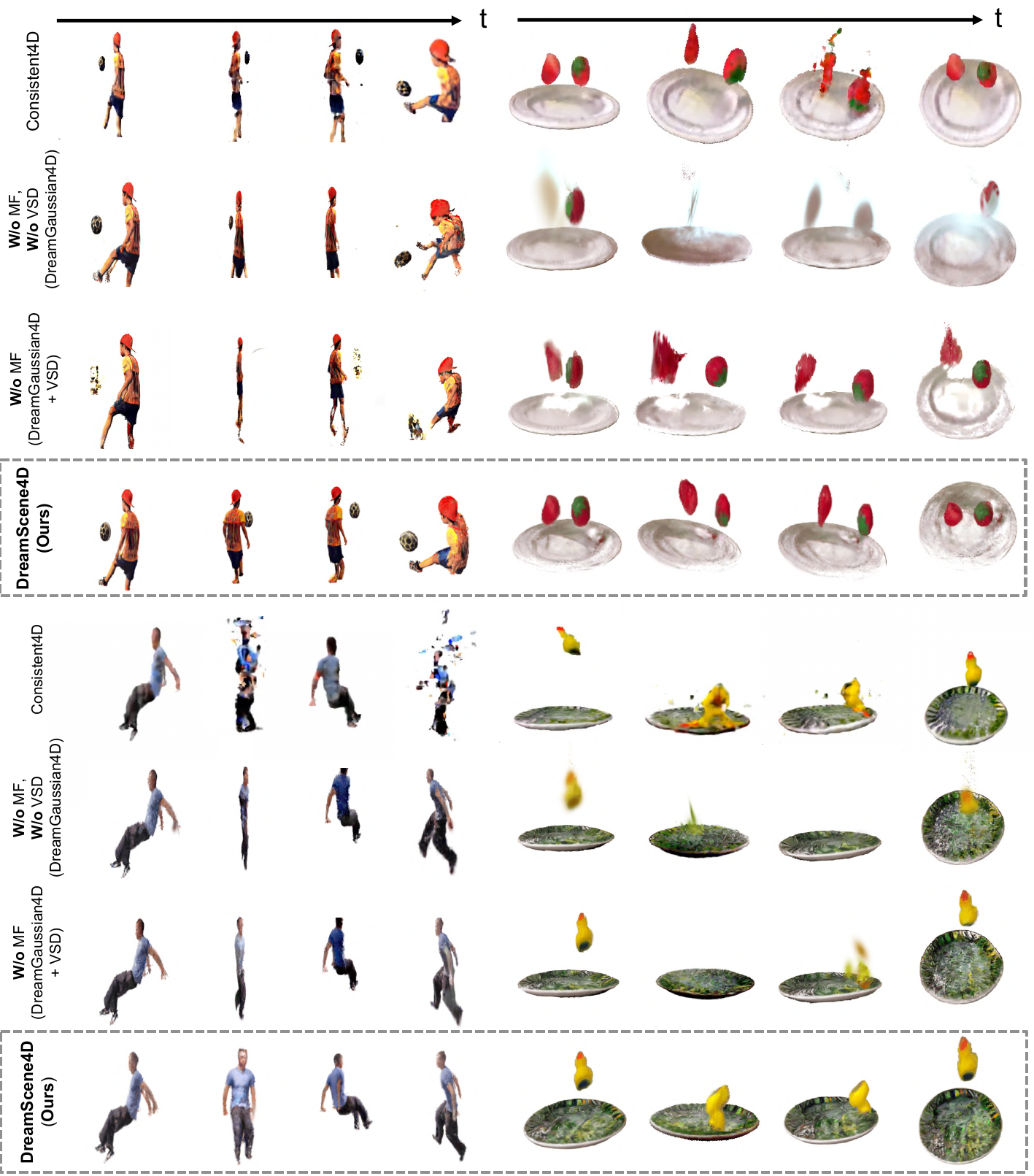}
    \caption{\textbf{Video to 4D Comparisons}. We render the Gaussians at various timesteps and camera views. We denote Motion Factorization as \textbf{MF} and Video Scene Decomposition as \textbf{VSD}. Our method produces consistent and faithful renders for fast-moving objects, while DreamGaussian4D~\cite{ren2023dreamgaussian4d} (2nd row) and Consistent4D~\cite{jiang2023consistent4d} (1st row) produce distorted 3D geometry, blurring, or broken artifacts. Refer to our Supp. Materials for extensive qualitative comparisons.}
    \label{fig:exp_vis}
    \vspace{-0.2in}
\end{figure}
\parsection{Datasets}
While there exist datasets used in previous video-to-4d generation works~\cite{jiang2023consistent4d}, they only consist of a small number of single-object synthetic videos with small amounts of motion. Thus, we evaluate the performance of \model{} on more challenging multi-object video datasets, including DAVIS~\cite{pont20172017},  Kubric~\cite{greff2022kubric}, and some self-captured videos with large object motion. 
We select a subset of 30 challenging real-world videos from DAVIS~\cite{pont20172017}, consisting of multi-object monocular videos with various amounts of motion. We further incorporate the labeled point trajectories from TAP-Vid-DAVIS~\cite{doersch2022tap} to evaluate the accuracy of the learned Gaussian deformations. In addition, we generated 50 multi-object videos from the Kubric~\cite{greff2022kubric} simulator, which provides challenging scenarios where objects can be small or off-center with fast motion.

\parsection{Evaluation Metrics} The quality of 4D generation can be measured in two aspects: the view rendering quality of the generated 3D geometry of the scene, and the accuracy of the 3D motion. For the former, we follow previous works~\cite{jiang2023consistent4d, ren2023dreamgaussian4d} and report the CLIP~\cite{radford2021learning} and LPIPS~\cite{zhang2018unreasonable} scores between 4 novel-view rendered frames and the reference frame, and compute its average score per video. These metrics allow us to assess the semantic similarity between rendered and reference frames. We also conducted a user study to evaluate the 4D generation quality for the DAVIS videos using two-way voting to compare each baseline with our method, where 50\% / 50\% indicates equal preference.

The accuracy of the estimated motion can be evaluated by measuring the End Point Error (EPE) of the projected 3D trajectories. For Kubric, we report the mean EPE separately for fully visible points and points that undergo occlusion. For DAVIS, we report the mean and median EPE~\cite{zheng2023pointodyssey}, as the annotations only exist for visible points.

\parsection{Implementation Details} We run our experiments on one 40GB A100 GPU. We crop and scale the individual objects to around 65\% of the image size for object lifting. For static 3D Gaussian optimization, we optimize for 1000 iterations with a batch size of 16. For optimizing the dynamic components, we optimize for 100 times the number of frames with a batch size of 10. More implementation and running time details are provided in the appendix.
\vspace{-0.1in}

\subsection{Video to 4D Scene Generation}
\label{sect:figures}

\parsection{Baselines} We consider the following baselines and ablated versions of our model: 

\noindent \textbf{(1)} Consistent4D~\cite{jiang2023consistent4d}, a recent state-of-the-art method for 4D generation from monocular videos that fits dynamic NeRFs per video using rendering losses and score distillation.

\noindent \textbf{(2)} DreamGaussian4D~\cite{ren2023dreamgaussian4d}, which uses dynamic 3D Gaussian Splatting like us for 4D generation from videos,  but does not use any video decomposition or motion factorization as \model{}. This is most related to our method. 

\noindent \textbf{(3)} DreamGaussian4D+VSD (Video Scene Decomposition). We augment DreamGaussian4D with VSD, where we segment every object before 4D lifting, and recompose them. The main difference between this stronger variant and our DreamScene4D is the lack of motion factorization.

\noindent \textbf{(4)} \model{} ablations on losses. We also ablate without flow losses and regularization losses.

\parsection{4D Generation Results on DAVIS \& Kubric} We present the 4D generation quality comparison in Table~\ref{tab:exp_vis}, where our proposed Video Scene Decomposition (VSD) and Motion Factorization (MF) schemes greatly improve the CLIP and LPIPS score compared to the input reference images. From the user study, we can also observe that DreamScene4D is generally preferred over each baseline.
Compared to the baselines, these significant improvements are mainly due to our proposed motion factorization, which enables the SDS loss to perform in an object-centric manner while reducing the training difficulty for the lightweight Gaussian deformation network in predicting large object motions.
We also show qualitative comparisons of 4D generation on multi-object videos and videos with large motion in Figure~\ref{fig:exp_vis}, where both variants of DreamGaussian4D~\cite{ren2023dreamgaussian4d} and Consistent4D~\cite{jiang2023consistent4d} tend to produce distorted 3D geometry, faulty motion, or broken artifacts of objects. 
This highlights the applicability of \model{} to handle real-world complex videos.

\parsection{4D Generation Results on Self-Captured Videos} 
We also captured some monocular videos with fast object motion using a smartphone to test the robustness of \model{}, where objects can be off-center and are subject to motion blur. We present qualitative results of the rendered 4D Gaussians in the right half of Figure \ref{fig:exp_vis}. Even under more casual video capturing settings with large motion blur, \model{} can still provide temporally consistent 4D scene generation results while the baselines generate blurry results or contain broken artifacts of the objects.
\vspace{-0.1in}

\begin{table}[!t]
 \vspace{-0.3in}
  \caption{\textbf{Video to 4D Scene Generation Comparisons}. We report the CLIP and LPIPS scores in Kubric~\cite{greff2022kubric} and DAVIS~\cite{pont20172017}. For user preference, A$\%$ / B$\%$ denotes that A$\%$ of the users prefer the \textit{baseline} while B$\%$ prefer \textit{ours} in two-way voting. We denote methods with Video Scene Decomposition as~\textbf{VSD} and methods with Motion Factorization as~\textbf{MF}.}
  \vspace{-0.05in}
  \label{tab:exp_vis}
  \centering
  \resizebox{1.0\linewidth}{!}{
  \begin{tabular}{lccccccc}
    \toprule
    \multirow{2}{*}{Method} & \multirow{2}{*}{\textbf{VSD}} & \multirow{2}{*}{\textbf{MF}} & \multicolumn{3}{c}{DAVIS} & \multicolumn{2}{c}{Kubric}\\
    \cmidrule(lr){4-6} \cmidrule(lr){7-8}
    & & & CLIP $\uparrow$ & LPIPS $\downarrow$ & User Pref. & CLIP $\uparrow$ & LPIPS $\downarrow$\\
    \midrule
    Consistent4D~\cite{jiang2023consistent4d} & - & - & 82.14 & \textbf{0.141} & 28.3\% / \textbf{71.7}\% & 80.46 & 0.117 \\
    DreamGaussian4D~\cite{ren2023dreamgaussian4d} & \xmark & \xmark & 77.81 & 0.181 & 22.1\% / \textbf{77.9}\% & 73.45 & 0.146 \\
    {DreamGaussian4D~\textbf{w/} VSD} & \cmark & \xmark & 81.39 & 0.169 & 30.4\% / \textbf{69.6}\% & 79.83 & 0.122 \\
    \midrule
    \model{} (\textbf{Ours}) & \cmark & \cmark & \textbf{85.09} & 0.152 & - & 85.53 & \textbf{0.112} \\
    { \textbf{w/o} $\mathcal{L}_{flow}$} & \cmark & \cmark & 84.94 & 0.152 & - & \textbf{86.41} & 0.113 \\
    { \textbf{w/o} $\mathcal{L}_{rigid}$ and $\mathcal{L}_{scale}$} & \cmark & \cmark & 83.24 & 0.153 & - & 84.07 & 0.115 \\
  \bottomrule
  \end{tabular}}
  \vspace{-0.25in}
\end{table}
\subsection{4D Gaussian Motion Accuracy}
\label{sect:4d_motion}
\parsection{Baselines and Ablations Design}  To evaluate the accuracy of the 4D Gaussian motion, we consider DreamGaussian4D~\cite{ren2023dreamgaussian4d} as the baseline, since extracting motion from NeRF-based methods~\cite{jiang2023consistent4d} is highly non-trivial. In addition, we compare against PIPS++~\cite{zheng2023pointodyssey} and CoTracker~\cite{karaev2023cotracker}, two fully-supervised methods explicitly trained for point-tracking, serving as upper bounds for performance.

\begin{figure}[!t]
    \centering
    \vspace{-0.05in}
    \includegraphics[width=1.0\textwidth]{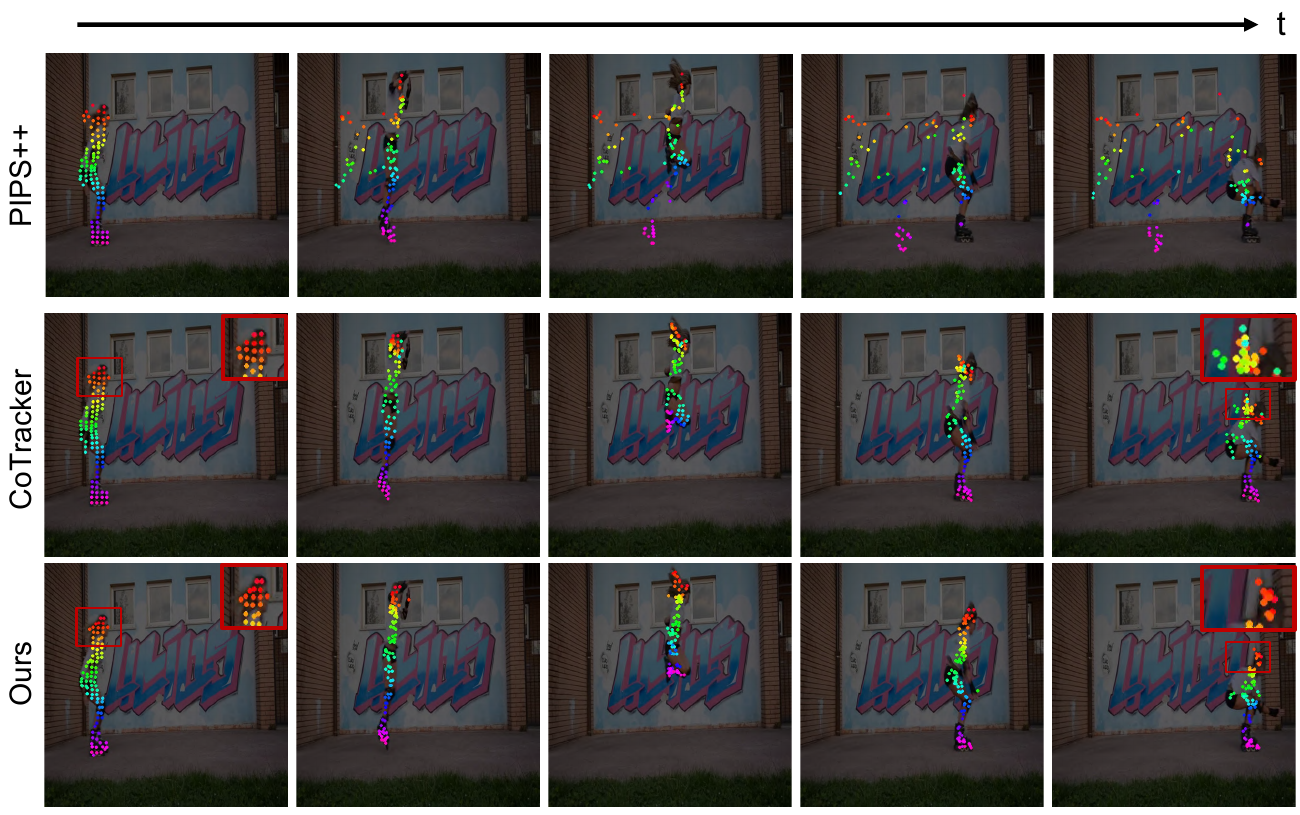}
    \vspace{-0.15in}
    \caption{\textbf{Motion Comparisons.} The 2D projected motion of Gaussians accurately aligns with dynamic human motion trajectory in the video, where the point trajectories estimated by PIPS++~\cite{zheng2023pointodyssey} tend to get ``stuck" in the background wall. For CoTracker~\cite{karaev2023cotracker}, partial point trajectories are mixed up, where some points in the chest region (\textcolor{Dandelion}{yellow}/\textcolor{ForestGreen}{green}) ending up in the head area (\textcolor{BrickRed}{red}).}
    \label{fig:exp_motion}
    \vspace{-0.15in}
\end{figure}

\parsection{4D Motion Accuracy in Video Reference Views} In Table~\ref{tab:exp_motion}, we tabulate the motion accuracy comparison, where \model{} achieves significantly lower EPE than the baseline DreamGaussian4D on both the DAVIS and Kubric datasets. 
We noted that conventional baselines often fail when objects are positioned near the edges of the video frame or undergo large motion. Interestingly, the motion accuracy of \model{} outperforms PIPS++~\cite{zheng2023pointodyssey}, despite never being trained on point tracking data, as in Figure~\ref{fig:exp_motion}. This is due to the strong object priors of \model{}, as the Gaussians adhere to remaining on the same object it generates and their motion is often strongly correlated.
\begin{table}[!t]
  \vspace{-0.18in}
  \caption{\textbf{Gaussian Motion Accuracy}. We report the EPE in Kubric~\cite{greff2022kubric} and DAVIS~\cite{pont20172017, doersch2022tap}. We denote methods with our Video Scene Decomposition in column \textbf{VSD} and methods with 3D Motion Factorization in column \textbf{MF}. Note that CoTracker is trained on Kubric.}
  \label{tab:exp_motion}
  \centering
  \resizebox{1.0\linewidth}{!}{
  \begin{tabu}{lcccccc}
    \toprule
    \multirow{2}{*}{Method} & \multirow{2}{*}{\textbf{VSD}} & \multirow{2}{*}{\textbf{MF}} & \multicolumn{2}{c}{DAVIS} & \multicolumn{2}{c}{Kubric} \\
    \cmidrule(lr){4-5} \cmidrule(lr){6-7}
    & & & EPE (vis) $\downarrow$ & EPE (occ) $\downarrow$ & Mean EPE $\downarrow$ & Median EPE $\downarrow$ \\
    \midrule
    \multicolumn{7}{l}{\textit{(a) \textbf{Not trained on point tracking data}}} \\
    \midrule
    Baseline: DreamGaussian4D~\cite{ren2023dreamgaussian4d} & \xmark & \xmark & 26.65 & 6.98 & 101.79 & 120.95  \\
    { \textbf{w/} VSD} & \cmark & \xmark & 20.95 & 6.72 & 85.27 & 92.42  \\
    \midrule
    \model{} (\textbf{Ours}) & \cmark & \cmark & \textbf{8.56} & 4.24 & \textbf{14.30} & \textbf{18.31}  \\
    { \textbf{w/o} $\mathcal{L}_{flow}$} & \cmark & \cmark & 10.91 & \textbf{3.83} & 18.54 & 24.51  \\
    { \textbf{w/o} $\mathcal{L}_{rigid}$ and $\mathcal{L}_{scale}$} & \cmark & \cmark & 10.29 & 4.78 & 16.21 & 22.29  \\
    \midrule
    \multicolumn{7}{l}{\textit{(b) \textbf{Trained on point tracking data}}} \\
    \midrule
    \rowfont{\color{gray}}
    PIPS++~\cite{zheng2023pointodyssey} & - & - & 19.61 & 5.36 & 16.72 & 29.65  \\
    \rowfont{\color{gray}}
    CoTracker~\cite{karaev2023cotracker} & - & - & 7.20 & 2.08 & 2.51 & 6.75  \\
  \bottomrule
  \end{tabu}}
  \vspace{-0.1in}
\end{table}

\parsection{4D Motion Results on Generated Novel Views} An advantage of representing the scene using 4D Gaussians is being able to obtain motion trajectories in arbitrary camera views, which we visualize in Figure~\ref{fig:teaser} and Figure~\ref{fig:exp_motion_multiview} in the appendix. \model{} can both generate a 4D scene with consistent appearance across views and produce temporally coherent motion trajectories in novel views.  

\subsection{Limitations}
Despite the exciting progress and results presented in the paper, several limitations still exist: \textbf{
(1)} The SDS prior fails to generalize to videos captured from a camera with steep elevation angles. \textbf{(2)} Scene composition may fall into local suboptimas if the rendered depth of the lifted 3D objects is not well aligned with the estimated depth. \textbf{(3)} Despite the inpainting, the Gaussians are still under-constrained when heavy occlusions happen, and artifacts may occur. \textbf{(4)} Our runtime scales linearly with the number of objects and can be slow for complex videos. Addressing these limitations by pursuing more data-driven ways for video to 4D generation is a direct avenue of our future work.

\section{Conclusion}
We presented \model{}, the first video-to-4D scene generation work to generate dynamic 3D scenes across occlusions, large object motions, and unseen viewpoints with both temporal and spatial consistency from multi-object monocular videos.
\model{} relies on decomposing the video scene into the background and individual object trajectories, and factorizes object motion to facilitate its estimation through pixel and motion rendering, even under large object displacements.
We tested \model{} on popular video datasets like DAVIS, Kubric, and challenging self-captured videos. 
\model{} infers not only accurate 3D point motion in the visible reference view but also provides robust motion tracks in synthesized novel views.  

\begin{ack}
This research was supported by Toyota Research Institute.
\end{ack}

\clearpage
{\small
    \bibliographystyle{ieee_fullname}
    \bibliography{main}
}



\clearpage
\appendix

\section{Appendix / Supplemental Material}


In the supplementary materials, we provide the details for video amodal completion, more implementation details of our DreamScene4D, and some qualitative and quantitative evaluations of the amodal completion. For more qualitative video-to-4D generation evaluations, we suggest looking at the videos in the \href{https://dreamscene4d.github.io/}{website}. 

\subsection{Video Amodal Completion}
We build off SD-Inpaint~\cite{Rombach2022stablediffusion} and adapt it for video amodal completion by making two modifications to the inference process without further fine-tuning.

\parsection{Spatial-Temporal Self-Attention}
\label{supp_:completion_st_att}
A common technique for extending Stable Diffusion-based models for video generation editing inflates the spatial self-attention layers to additionally attend across frames without changing the pre-trained weights~\cite{wu2023tune, khachatryan2023text2video, ceylan2023pix2video, qi2023fatezero}. Similar to~\cite{ceylan2023pix2video}, we inject tokens from adjacent frames during self-attention to enhance inpainting consistency. Specifically, the self-attention operation can be denoted as:
\begin{equation}
    Q = W_Q z_t,\, K = W_K \left[ z_{t-1}, z_t, z_{t+1} \right],\, V = W_V \left[ z_{t-1}, z_t, z_{t+1} \right],
    \label{eq:supp_st_att}
\end{equation}
where $\left[ \cdot \right]$ represents concatenation, $z_t$ is the latent representation of frame $t$, and $W_Q$, $W_K$, and $W_V$ denote the (frozen) projection matrices that project inputs to queries, keys, and values.

\parsection{Latent Consistency Guidance}
\label{supp:latent_guidance}
While inflating the self-attention layers allows the diffusion model to attend to and denoise multiple frames simultaneously, it does not ensure that the inpainted video frames are temporally consistent. 
To solve this issue, we take inspiration from previous works that perform test-time optimization while denoising for structured image editing~\cite{parmar2023zero} and panorama generation~\cite{lee2024syncdiffusion} and explicitly enforce the latents during denoising to be consistent.

Concretely, we follow a two-step process for each denoising step for noisy latent $z^{\tau}$ at denoising timestep $\tau$ to latent $z^{\tau-1}$. For each noisy latent $z^{\tau}_t$ at frame $t$, we compute the fully denoised latent $z^0_t$ and its corresponding image $\hat{I}_t$ directly in one step. To encourage the latents of multiple frames to become semantically similar, we freeze the network and only update $z^{\tau}$:
\begin{equation}
    \hat{z}^{\tau} = z^{\tau} - \eta \nabla_{z} \mathcal{L}_c,
    \label{eq:supp_guidance_step}
\end{equation}
where $\eta$ determines the size of the gradient step and $\mathcal{L}_c$ is a similarity loss, i.e., CLIP feature loss or the SSIM between pairs of $\hat{I}_t$. 
After this latent optimization step, we take $\hat{z}^{\tau}$ and predict the added noise $\hat{\epsilon}^{\tau}$ using the diffusion model to compute $z^{\tau-1}$ as:
\begin{equation}
    z^{\tau-1}=\sqrt{\alpha_{t-1}} \left(\frac{\hat{z}^{\tau}-\sqrt{1-\alpha_t} \hat{\epsilon}^{\tau}}{\sqrt{\alpha_t}}\right) + \sqrt{1-\alpha_{t-1}} \hat{\epsilon}^{\tau},
    \label{eq:supp_guidance_denoise}
\end{equation}
where $\alpha_t$ is the noise scaling factor defined in DDIM~\cite{song2020denoising}.

\subsection{More Implementation Details}

\parsection{Deformation Network.} The deformation network uses a Hexplane~\cite{cao2023hexplane} backbone representation with a 2-layer MLP head on top to predict the required outputs. In our evaluations, the resolution of the Hexplanes is $[64, 64, 64, 25]$ for $(x, y, z, t)$ to ensure fair comparisons with the baselines. For longer videos (more than 32 frames), we set the resolution to $[64, 64, 64, 0.8T]$ for $(x, y, z, t)$, where $T$ is the number of frames. We found that the network is generally quite robust to the temporal resolution of the Hexplane grid.

\parsection{Learning Rate.} Following DreamGaussian~\cite{tang2023dreamgaussian} and DreamGaussian4D~\cite{ren2023dreamgaussian4d}, we set different learning rates for different Gaussian parameters. We use the same set of hyperparameters as DreamGaussian and use a learning rate that decays from $1e^{-3}$ to $2e^{-5}$ for the position, a static learning rate of $0.01$ for the spherical harmonics, $0.05$ for the opacity, ad $5e^{-3}$ for the scale and rotation. The learning rate of the Hexplane grid is set to $6.4e^{-4}$ while the learning rate of the MLP prediction heads is set to $6.4e^{-3}$. During joint fine-tuning of the deformation network and the object-centric to world frame transformations, we set the learning rate to 0.1x the original value. We use the AdamW optimizer for all our optimization processes.

\parsection{Densification and Pruning.} Following \cite{tang2023dreamgaussian,ren2023dreamgaussian4d}, the densification in the image-to-3D step is applied for Gaussians with accumulated gradient larger than $0.5$ and max scaling smaller than $0.05$. Gaussians with an opacity value less than $0.01$ or max scaling larger than $0.05$ are also pruned. This is done every $100$ optimization step. Densification and pruning are both disabled during motion optimization.

\parsection{Running Time.} As mentioned in the main text, we perform 1000 optimization steps for the static 3D Gaussian splatting process, while the deformation optimization takes $100 \cdot T$ optimization steps, where $T$ is the number of frames. The joint fine-tuning process is conducted over $100$ steps. While many videos converge faster, we found that videos with more complex objects and motion require more optimization steps. On a 40GB A100 GPU, the static 3D lifting process takes around 5.5 minutes, and the 4D lifting process takes around 17 minutes for a video of 16 frames per object.

\parsection{Evaluation Settings.} In our video-to-4D evaluations, we render from the following combination of (elevation, azimuth) angles: (0, 45), (0, -45), (45, 0), (-45, 0). These novel view renders are then compared with the reference view at each timestep to obtain the CLIP and LPIPS scores. The scores are then averaged across all views and timesteps for the final score.

\parsection{User Preference Study.} For the user study, we take the 30 DAVIS videos and produce a smooth orbital render video by varying the azimuth angle while rendering the deforming object(s). We use Amazon Turk to outsource evaluations on the 30 DAVIS videos for each baseline, including DreamGaussian4D~\cite{ren2023dreamgaussian4d}, DreamGaussian4D~\cite{ren2023dreamgaussian4d} + Video Scene Decomposition (VSD), Consistent4D~\cite{jiang2023consistent4d} and our DreamScene4D. Each set of videos is reviewed by 30 workers with a HIT rate of over 95\% for a total of 2700 answers collected. 
The whole user preference study takes about 97s per question and 72.8h working hours in total.
We manually filtered out workers who submitted the same answer for all the videos and assigned new ones during the collection process until the desired number of answers had been collected.

\begin{figure}[!t]
    \centering
    \includegraphics[width=1.0\textwidth]{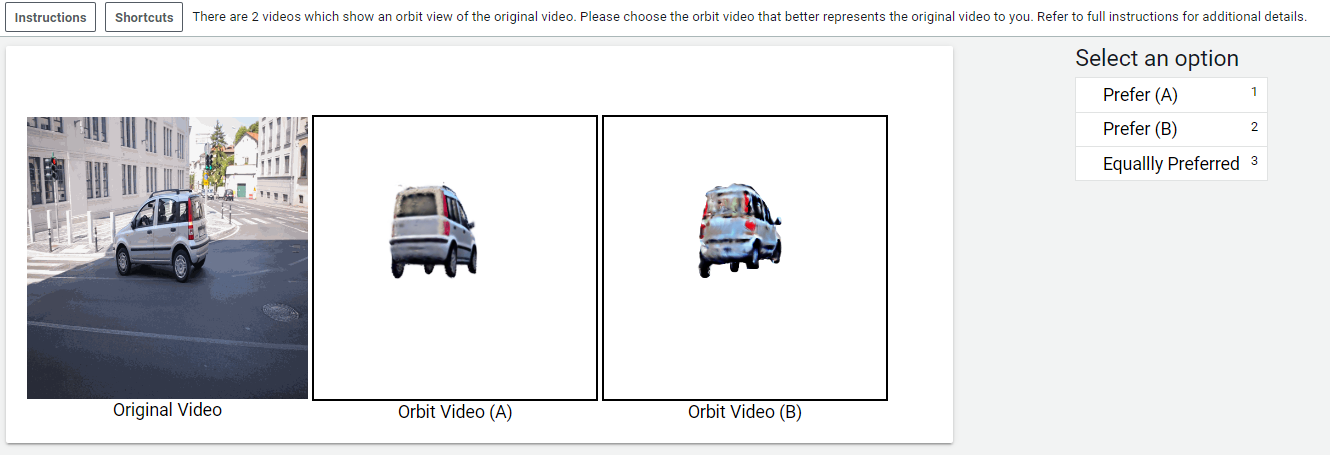}
    \caption{\textbf{User survey interface.} A GUI example of what an Amazon Turk worker would see as part of the user preference study.}
    \label{fig:supp_survey}
\end{figure}

The full instruction given is as follows:

\begin{quote}
\textit{Please read the instructions and check the videos carefully.}
\bigskip

\textit{There are 2 videos that show an orbit view of the original video. Please choose the orbiting video that looks more realistic and better represents the original video to you. The options (A) and (B) correspond to the two given orbit videos. If you think that both are of the same quality, please select Equally Preferred.}

\bigskip
\textit{To judge the quality of the videos, consider the following points:}

\bigskip
\textit{1. Do the objects in the orbit video correspond to the original video?}

\textit{2. Does the video look geometrically correct (e.g. not overly flat) when the camera is orbiting?}

\textit{3. Are there any visual artifacts (e.g. floaters, weird textures) during the orbit?}

\bigskip
\textit{\textbf{Please ignore the background in the original video.}}
\end{quote}

A GUI sample of a survey question is also provided in Figure~\ref{fig:supp_survey} for reference.

\subsection{Additional Results}

\begin{figure}[!t]
    \centering
    \includegraphics[width=1.0\textwidth]{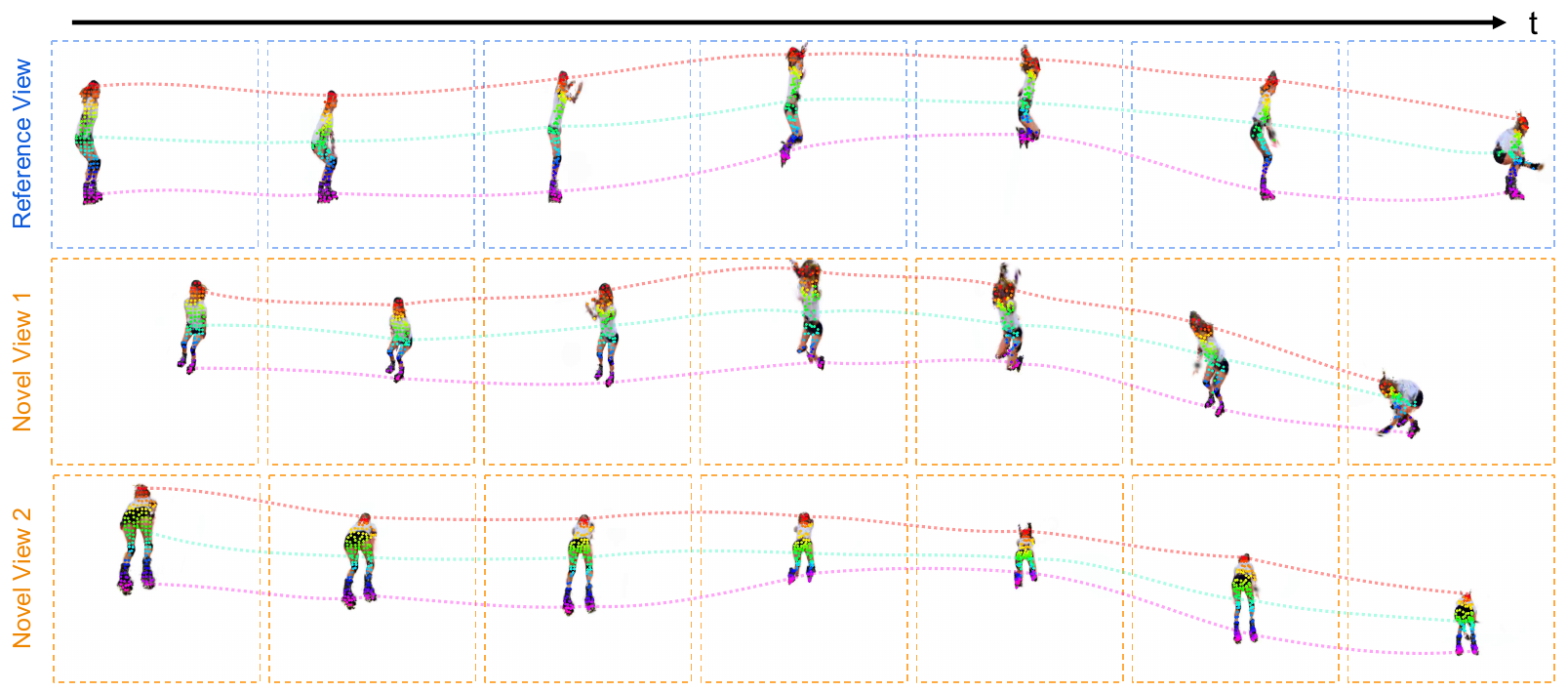}
    \vspace{-0.25in}    \caption{\textbf{Gaussian Motion Visualizations.} We visualize the Gaussian trajectories in the reference view corresponding to the video as well as in multiple novel views. The rendered Gaussians are sampled \textit{independently} for each view. \model{} can produce accurate motion in different camera poses \textbf{w/o} explicit point trajectory supervision.}
\label{fig:exp_motion_multiview}
    \vspace{-0.2in}
\end{figure}

\parsection{4D Motion Visualizations in Novel Views} Since \model{} represents the scene using 4D Gaussians, it is able to obtain motion trajectories in arbitrary camera views, as in Figure~\ref{fig:exp_motion_multiview}. \model{} can both generate a 4D scene with consistent appearance across views and produce temporally coherent motion trajectories.


\begin{figure}[!ht]
    \centering
    \includegraphics[width=1.0\textwidth]{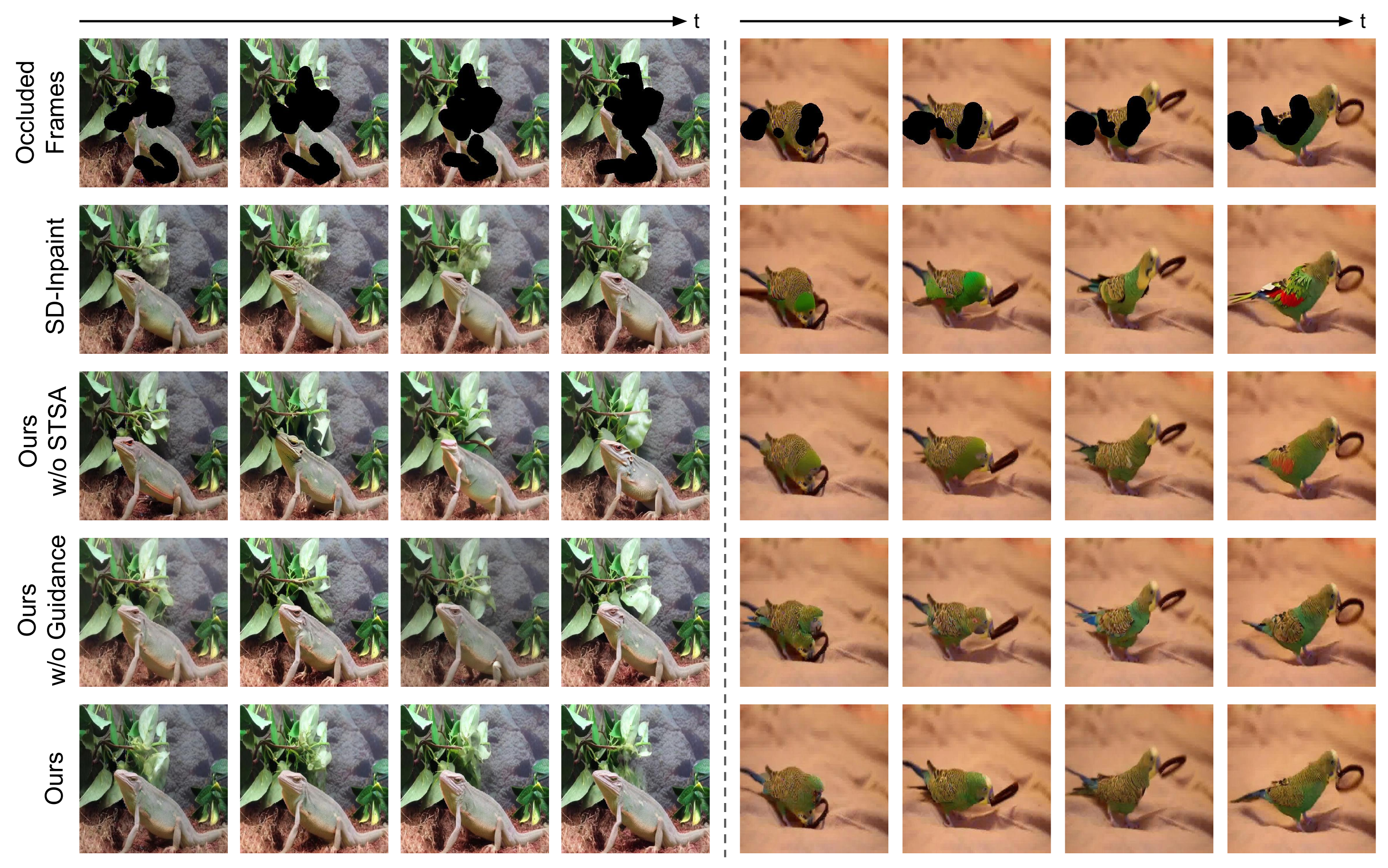}
    \caption{\textbf{Video Amodal Completion Comparisons.} Spatiotemporal self-attention and Consistency Guidance both help to preserve the identity consistency of the inpainted objects.}
    \label{fig:amodal_completion}
\end{figure}

\parsection{Video Amodal Completion} To ablate our extensions to SD-Inpaint for video amodal completion, we randomly select 120 videos from YoutubeVOS~\cite{xu2018youtube} and generate random occlusion masks in the video~\cite{suvorov2022resolution, chang2019learnable}. We compare against Repaint~\cite{lugmayr2022repaint} and SD-Inpaint~\cite{Rombach2022stablediffusion} for video amodal completion. Both baseline methods are based on Stable Diffusion~\cite{Rombach2022stablediffusion}. Repaint alters the reverse diffusion iterations by sampling the unmasked regions of the image. SD-Inpaint, on the other hand, finetunes Stable Diffusion for free-form inpainting. We also ablate the performance of our proposed amodal completion approach without the inflated spatiotemporal self-attention (denoted as STSA) and consistency guidance.
We summarize the results in Table~\ref{tab:supp_amodal_completion} and show some visual comparisons in Figure~\ref{fig:amodal_completion}.
Our modification achieves more consistent and accurate video completion than image inpainting approaches by leveraging temporal information during the denoising process. Note that these techniques complement other video completion approaches since \model{} mainly focuses on video-to-4D scene generation.

\begin{table}[th]
  \caption{\textbf{Video Amodal Completion Evaluations.} We report the PSNR, LPIPS, and Temporal Consistency (TC) measured using CLIP similarity in randomly masked YoutubeVOS~\cite{xu2018youtube} videos.}
  \label{tab:supp_amodal_completion}
  \centering
  \begin{tabular}{lcccc}
    \toprule
    Method & PSNR $\uparrow$ & \makecell{PSNR $\uparrow$\\(masked)} & LPIPS $\downarrow$ & TC $\uparrow$ \\
    \midrule
    Repaint~\cite{lugmayr2022repaint} & 20.76 & \makecell{14.04} & 0.23 & 91.18 \\
    SD-Inpaint~\cite{Rombach2022stablediffusion}  & 21.07 & \makecell{14.35} & 0.23 & 91.72 \\
    \midrule
    \model{} (\textbf{Ours}) & \textbf{22.27} & \makecell{\textbf{16.09}} & \textbf{0.22} & \textbf{93.40} \\
    \textbf{w/o} STSA  & 21.56 & \makecell{15.31} & 0.23 & 92.58 \\
    \textbf{w/o} Guidance  & 21.71 & \makecell{15.20} & 0.23 & 92.91 \\
  \bottomrule
  \end{tabular}
\end{table}

\begin{figure}[!t]
    \centering
    \includegraphics[width=1.0\textwidth]{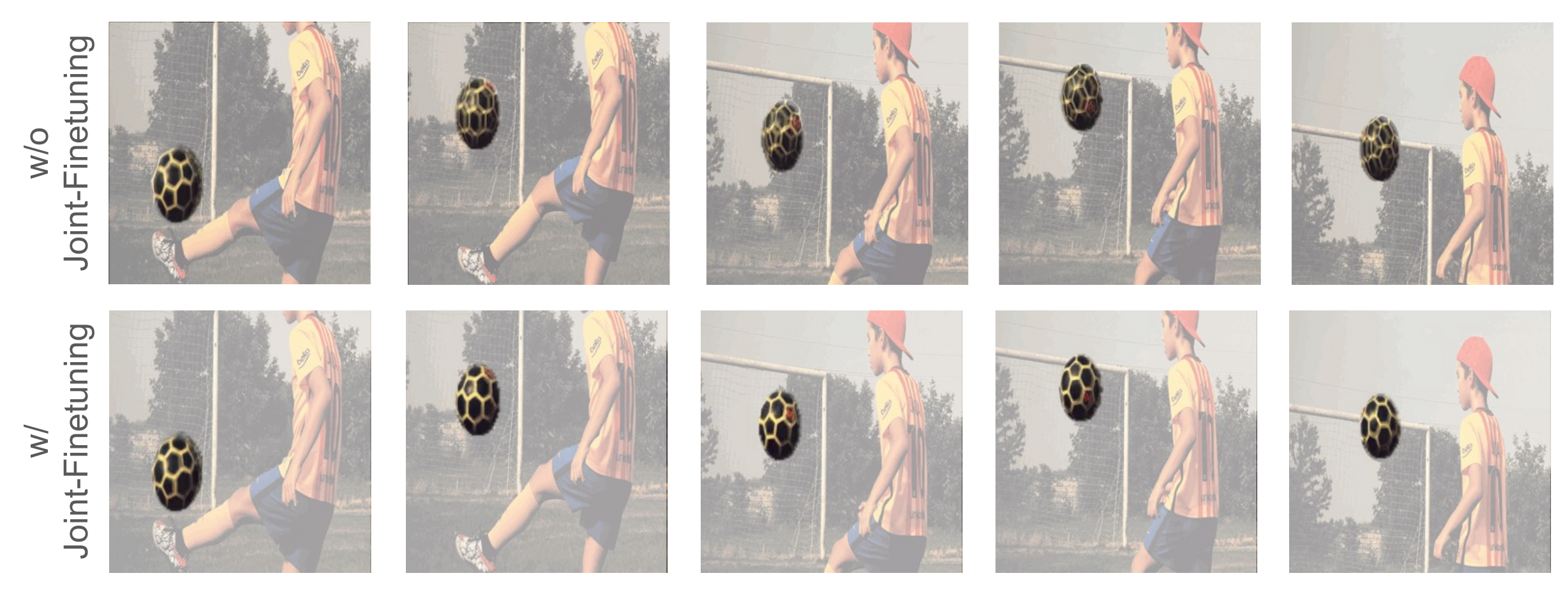}
    \caption{\textbf{Mitigating the parallax effect.} A small amount of joint fine-tuning steps can help mitigate the parallax effect and align the rendered Gaussians to the input video frames.}
    \label{fig:supp_parallax}
\end{figure}

\parsection{Mitigating Parallax Effects via Joint Optimization} We show an example of the rendered Gaussians before and after performing the joint optimization for the deformation network and the object-centric to world frame transformations in Figure~\ref{fig:supp_parallax}. We can see that a small amount of joint fine-tuning steps helps alleviate the parallax effect and better aligns the rendered Gaussians to the input video frames.

\subsection{Failure Cases}
\begin{figure}[!ht]
    \centering
    \includegraphics[width=1.0\textwidth]{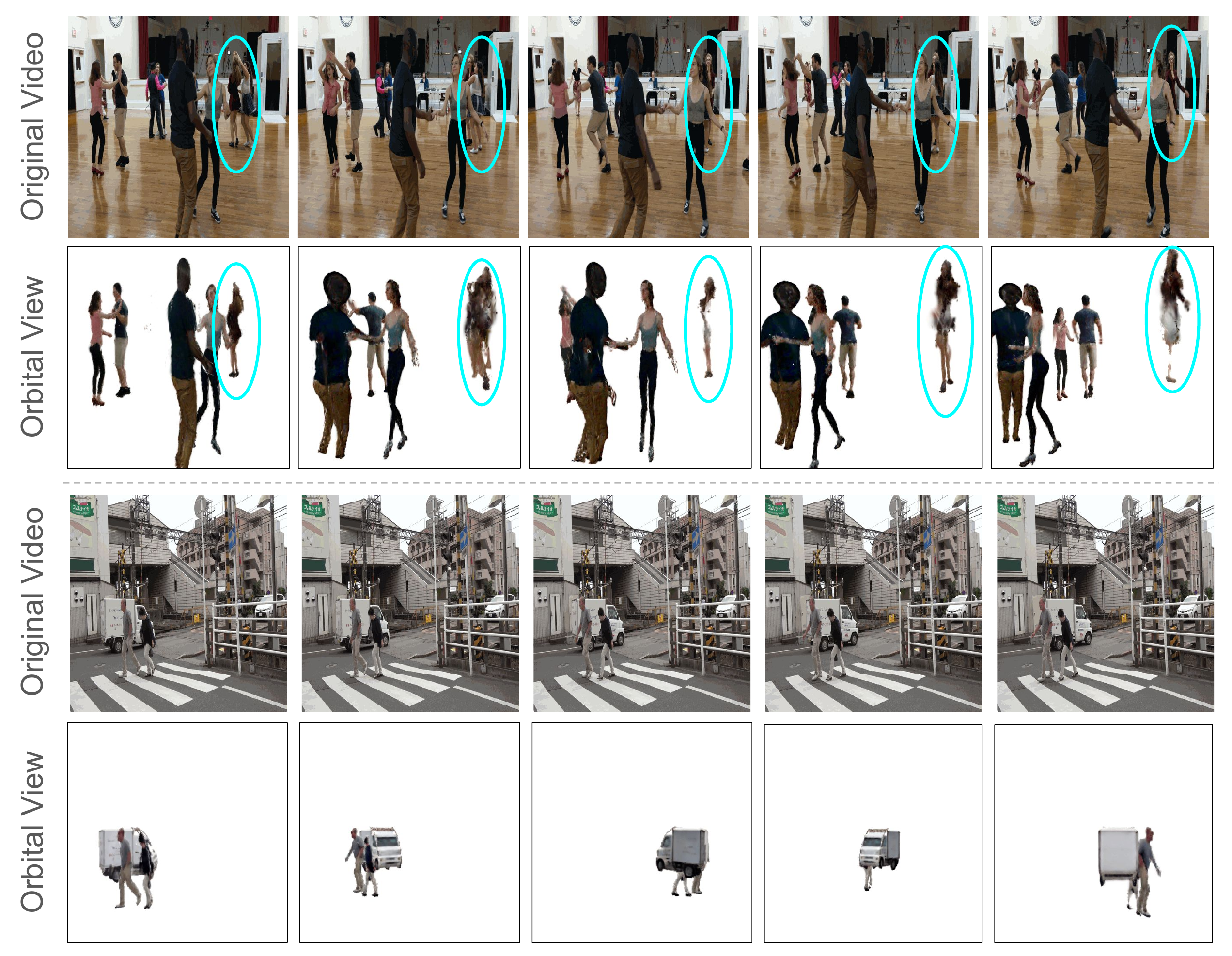}
    \caption{\textbf{Failure Cases.} We show 2 representative failure cases. The first case (top 2 rows) is due to inpainting failures (circled in blue), where the inpainted frames are not of high quality, leading to flickering objects when rendered. The second case (bottom 2 rows) arises from poor depth predictions, which leads to composition errors. The two humans are placed too close to the truck, making the scale proportions of the objects seem unnatural (i.e. the truck is too small). }
    \label{fig:failure}
\end{figure}

We additionally show some failure cases corresponding to the limitations documented in the main text in Figure~\ref{fig:failure}. Based on our observations, the inpainting is very unstable during heavy occlusions. We believe that instead of solely relying on rendering losses for the occluded regions, incorporating some form of semantic guidance loss (e.g. CLIP feature loss) might be a promising direction.

\subsection{Broader Impact}
Our approach is deeply connected to VR/AR applications and can potentially provide 3D meshes and dense 3D trajectories for robot manipulation. While our method does not generate or modify the original video, it is still possible for users to use generative models with malicious intent, and then apply our approach for video-to-4D lifting. The potential negative impact can be avoided by applying preventative measures in generative models and rejecting the video input if violations are found.

\subsection{DAVIS Split}
We list the DAVIS video names that were used to perform evaluations:
\begin{verbatim*}
bear,blackswan,bmx-bumps,boxing-fisheye,car-shadow,cows,crossing,
dance-twirl,dancing,dog-gooses,dogs-jump,gold-fish,hike,hockey,kid-football,
lab-coat,lindy-hop,longboard,lucia,night-race,parkour,pigs,rallye,rhino,
rollerblade,schoolgirls,scooter-black,scooter-gray,snowboard,stroller,train
\end{verbatim*}

For bmx-bumps, longboard, scooter-black, and scooter-gray, we merge the mask of the human and the other objects into one as they move together for the entire video (e.g. person riding a bike or a scooter).


\end{document}